%% file: main.tex
\documentclass[10pt, a4paper]{article}

\usepackage{lrec-coling2024} 

\usepackage[utf8]{inputenc} 
\usepackage[T1]{fontenc}    
\usepackage{hyperref}       
\usepackage{url}            
\usepackage{booktabs}       
\usepackage{amsfonts}       
\usepackage{nicefrac}       
\usepackage{microtype}      
\usepackage{xcolor}         
\usepackage{times}
\usepackage{epsfig}
\usepackage{graphicx}
\usepackage{amsmath}
\usepackage{amssymb}
\usepackage{multirow}
\usepackage{subcaption}
\usepackage{CJKutf8}
\usepackage[normalem]{ulem}
\usepackage{wrapfig}
\usepackage[inline]{enumitem}
\usepackage{pifont}

\usepackage{algorithm}
\usepackage{algorithmicx}
\usepackage{algpseudocode}

\usepackage[capitalize]{cleveref}
\crefname{section}{Sec.}{Secs.}
\Crefname{section}{Section}{Sections}
\Crefname{table}{Table}{Tables}
\crefname{table}{Tab.}{Tabs.}

\newcommand{\sub}[1]{\textcolor{red}{#1}}
\newcommand{\del}[1]{\textcolor{blue}{#1}}
\newcommand{\ins}[1]{\textcolor{green}{#1}}

\newcommand{\cmark}{\ding{51}}%

\newcommand{\numhours}{16.07}
\newcommand{\numglosses}{6,515}
\newcommand{\numchars}{2,850} 
\newcommand{\numhourssa}{11.66}
\newcommand{\numhourssb}{4.41}
\newcommand{\numsamplessa}{5,350}

\newcommand{\bestwertest}{34.08\%}
\newcommand{\bestbleutest}{23.58}

\title{A Hong Kong Sign Language Corpus Collected from Sign-interpreted TV News}

\name{
	Zhe Niu\textsuperscript{\rm 1,*}\thanks{* Equal contribution.} \quad Ronglai Zuo\textsuperscript{\rm 1,*}\footnotemark[1] \quad Brian Mak\textsuperscript{\rm 1} \quad Fangyun Wei\textsuperscript{\rm 2}\\
}

\address{
	\textsuperscript{\rm 1}The Hong Kong University of Science and Technology \quad \textsuperscript{\rm 2}Microsoft Research Asia\\
	\texttt{\{zniu,rzuo,mak\}@cse.ust.hk} \quad \texttt{fawe@microsoft.com} \\
}

\begin{document}

\input{sections/abs}

\maketitleabstract

\input{sections/intro}
\input{sections/related}
\input{sections/stats}
\input{sections/method}
\input{sections/expr_v5.8.tex}

\input{sections/conclusion}

\nocite{*}
\section{Bibliographical References}\label{sec:reference}

\bibliographystyle{lrec-coling2024-natbib}
\bibliography{main}

\end{document}

%% file: sections/abs.tex
\abstract
{
	This paper introduces TVB-HKSL-News, a new Hong Kong sign language (HKSL) dataset collected from a TV news program over a period of 7 months. The dataset is collected to enrich resources for HKSL and support research in large-vocabulary continuous sign language recognition (SLR) and translation (SLT). It consists of \numhours{} hours of sign videos of two signers with a vocabulary of \numglosses{} glosses (for SLR) and \numchars{} Chinese characters or 18K Chinese words (for SLT). One signer has \numhourssa{} hours of sign videos and the other has \numhourssb{} hours. One objective in building the dataset is to support the investigation of
	how well large-vocabulary continuous sign language recognition/translation can be done for a single signer given a (relatively) large amount of his/her training data, which could potentially lead to the development of new modeling methods.
	Besides, most parts of the data collection pipeline are automated with little human intervention; we believe that our collection method can be scaled up to collect more sign language data easily for SLT in the future for any sign languages if such sign-interpreted videos are available.
	We also run a SOTA SLR/SLT model on the dataset and get a baseline SLR word error rate of \bestwertest{} and a baseline SLT BLEU-4 score of \bestbleutest{} for benchmarking future research on the dataset.
	\\ \newline
	\Keywords{Sign Language Dataset, Sign Language Recognition (SLR), Sign Language Translation (SLT).}
}

%% file: sections/intro.tex
\section{Introduction}

Sign language (SL) is the primary mode of communication for many individuals who are deaf or hard of hearing. Advances in large deep networks demand more and larger datasets to support the development and research of sign language recognition (SLR) and sign language translation (SLT).
Consequently, several large-scale datasets~\cite{phoenix14,phoenix14t,csl100,csl-daily,bobsl} have been created for many sign languages such as ASL, BSL, CSL, DGS, etc.
Given that there are currently few publicly available resources for Hong Kong Sign Language (HKSL) research, we introduce, in this paper, a new large-vocabulary HKSL dataset, called TVB-HKSL-News, for the development of SLR and SLT for HKSL.

\begin{figure}[t]
	\centering
	\includegraphics[width=\linewidth]{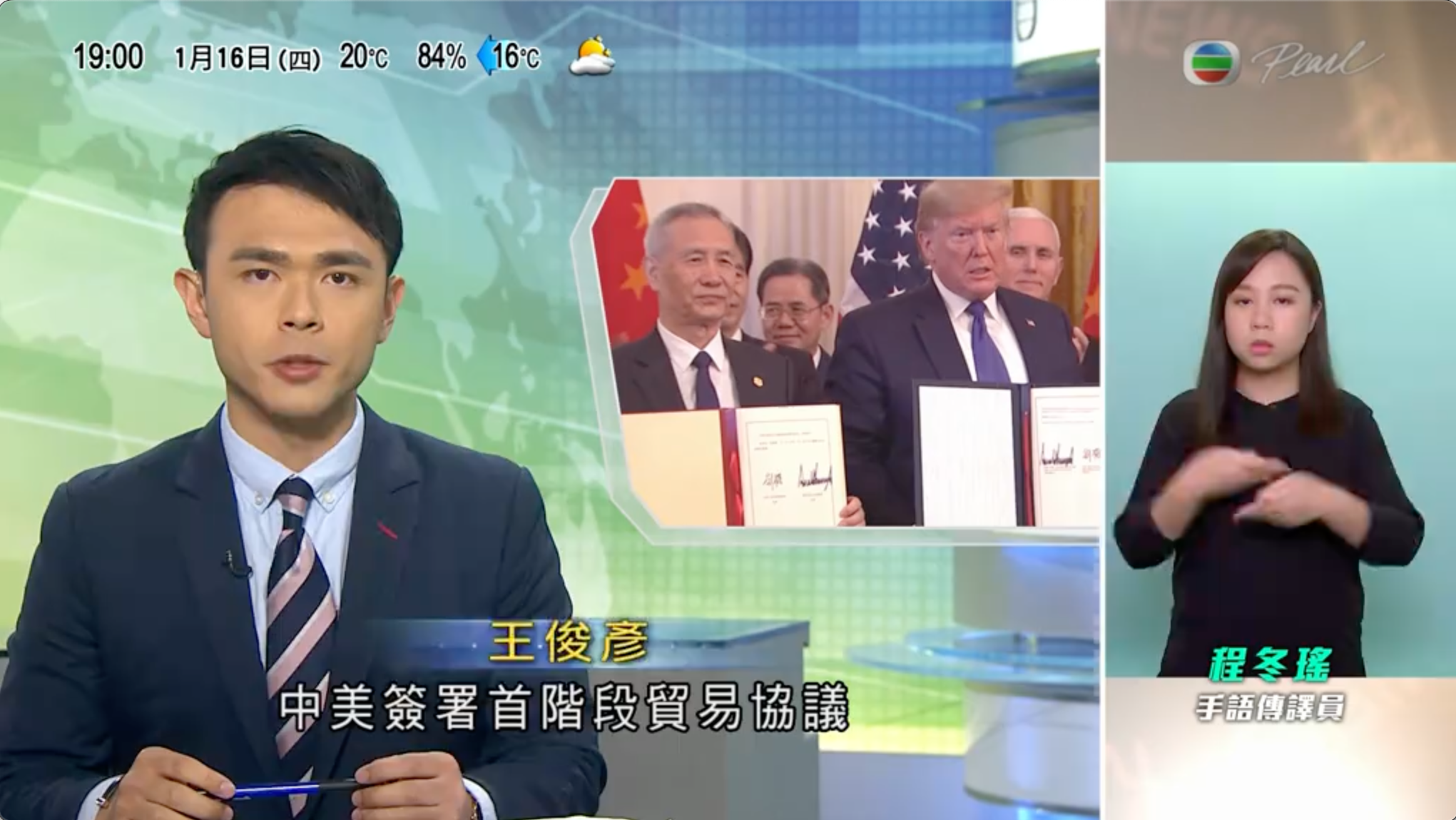}
	\caption{In the "TVB News Report with Sign Language" program, a news anchor speaks on the left side of the screen while an HKSL interpreter signs on the right. Chinese subtitles are displayed at the bottom of the screen.}
	\label{fig:hksl-raw}
\end{figure}

The TVB-HKSL-News dataset contains HKSL videos collected from the News Report with Sign Language program (as depicted in~\Cref{fig:hksl-raw}) from the Television Broadcasts Limited (TVB) over a period of seven months.
The dataset is intended to support research in large-vocabulary, signer-dependent continuous SLR and SLT. It includes \numhours{} hours of sign videos from two signers, with a vocabulary of \numglosses{} glosses (for SLR) and \numchars{}/18K Chinese characters/words (for SLT).
One primary objective of building the dataset is to investigate new modeling methods for how well large-vocabulary SLR/SLT can be done for a single signer, given a relatively large amount of his/her training data. This is lacking in current HKSL resources and is not common even in other sign languages (with the exception of BOBSL \cite{bobsl}). With a larger amount of data available for a signer, researchers can explore more sophisticated modeling techniques and evaluate the impact of amount of data on SLR and SLT.

\begin{table*}[t]
	\caption{Comparison of some existing datasets for continuous SLR and SLT with TVB-HKSL-News. The superscript $\dagger$ represents character count rather than word count, which is more suitable for gauging translation performance in Chinese~\cite{why-slt-character}. The approximate word counts for CSL-Daily and TVB-HKSL-News are 8K and 18K, respectively. Our word counts are estimated using Jieba~\cite{jieba}.}
	\label{table:related}
	\centering
	\resizebox{1.0\linewidth}{!}{
		\begin{tabular}{@{}c|cccccccccc@{}}
			\toprule
			Dataset/Work                     & Lang. & SLR Vocab.  & SLT Vocab.              & Duration (h) & Resolution       & FPS & \#Videos & \#Signers & Source \\
			\midrule
			GSL~\cite{gsl}                   & GSL   & 310         & -                       & 9.59         & 848$\times$480   & 30  & 10K      & 7         & Lab    \\
			PHOENIX-14~\cite{phoenix14}      & DGS   & 1,081       & -                       & 12.54        & 210$\times$260   & 25  & 7K       & 9         & TV     \\
			PHOENIX-14T~\cite{phoenix14t}    & DGS   & 1,066       & 2,887                   & 10.53        & 210$\times$260   & 25  & 8K       & 9         & TV     \\
			CSL100~\cite{csl100}             & CSL   & 178         & -                       & 100          & 1280$\times$720  & 30  & 25K      & 50        & Lab    \\
			CSL-Daily~\cite{csl-daily}       & CSL   & 2,000       & 2,370$^\dagger$         & 23.27        & 1920$\times$1080 & 30  & 21K      & 10        & Lab    \\
			BOBSL~\cite{bobsl}               & BSL   & 2,281       & 78K                     & 1,467        & 444$\times$444   & 25  & 1.2M     & 39        & TV     \\
			SignBERT~\cite{zhou2021signbert} & HKSL  & 55          & -                       & -            & 1920$\times$1080 & 30  & 2K       & 6         & Lab    \\
			\midrule
			{TVB-HKSL-News (ours)}           & HKSL  & \numglosses & \numchars\!\!$^\dagger$ & ~\numhours   & 248$\times$360   & 25  & 7K       & 2         & TV     \\
			\bottomrule
		\end{tabular}
	}
\end{table*}

Besides the TVB-HKSL-News dataset, we also introduce a data collection pipeline for SL data from TV programs, which was utilized during the collection of our dataset. This pipeline includes an automatic collection process for Sign Language Translation (SLT) data and a computer-assisted annotation process for Sign Language Recognition (SLR) data.
In the initial phase, we collect SLT data from the TVB program, and process them to obtain sign video segments and their corresponding text transcriptions. This stage leverages various computer vision techniques to perform detection of sign activities, extraction of subtitles, and alignment between sign and subtitle segments.
We then engage professional sign language annotators and design an annotation software for them to label the sign glosses in each collected sign video segment for SLR research. The software presents sign video segments alongside the text transcriptions from the previous stage, providing a convenient tool for efficient sign language glossing.

Following the compilation of our dataset, we conduct experiments with various state-of-the-art models~\cite{s3d,chentwo,zuo2022c2slr} for both SLR and SLT.
We establish a strong baseline WER of \bestwertest{} and a baseline BLEU-4 score of \bestbleutest{}, respectively. These results serve as benchmarks for future research on the TVB-HKSL-News dataset.
Furthermore, we also investigate signer-dependent tasks, and explore the impact of training dataset size under this setting. We demonstrate how the amount of training data affects performance and our findings indicate that additional data from a different signer can also lead to small improvements.
In summary, we present the following contributions:
\begin{itemize}[leftmargin=*]
	\item We publish a new HKSL dataset called TVB-HKSL-news to support large-vocabulary continuous SLR/SLT for HKSL.
	\item We present an automated pipeline for collecting substantial SL data from subtitled, sign-interpreted videos for SLT research, along with an annotation software to facilitate sign glossing for SLR studies.
	\item We establish an SLR/SLT benchmark on the new dataset using a state-of-the-art SLR/SLT model.
\end{itemize}
Instructions to access the dataset can be found on our dataset webpage\footnote{Dataset webpage: \url{https://tvb-hksl-news.github.io/}}.

%% file: sections/related.tex
\section{Related Works}

\begin{table*}[t]
	\caption{Statistics of the TVB-HKSL-News train/development/test data split.}
	\label{table:hksl-stats}
	\centering
	\resizebox{0.85\linewidth}{!}{
		\begin{tabular}{@{}c|cc|cccc|cccc@{}}
			\toprule
			        & \multirow{2}{*}{Hours} & \multirow{2}{*}{\# Samples} & \multicolumn{4}{|c|}{Glosses for SLR} & \multicolumn{4}{c}{Chinese Characters for SLT}                                                                       \\
			        &                        &                             & Vocab                                 & Running                                        & \# OOVs & \# Singletons & Vocab & Running & \# OOVs & \# Singletons \\ \midrule
			Train   & 14.71                  & 6,516                       & 6,515                                 & 111,204                                        & N/A     & 2,925         & 2,816 & 212,108 & N/A     & 466           \\
			Dev     & 0.67                   & 322                         & 1,091                                 & 5,222                                          & 0       & 471           & 1,279 & 10,003  & 17      & 395           \\
			Test    & 0.69                   & 322                         & 1,130                                 & 5,391                                          & 0       & 518           & 1,276 & 10,199  & 19      & 399           \\
			\midrule
			Overall & 16.07                  & 7,160                       & 6,515                                 & 121,817                                        & N/A     & 2,820         & 2,850 & 232,310 & N/A     & 462           \\
			\bottomrule
		\end{tabular}
	}
\end{table*}

Numerous sign language (SL) datasets have been created for various tasks, including SL spotting~\cite{s-pot}, SLR~\cite{gsl, csl100, phoenix14, phoenix14t, bosphorus-sign-22k, wlasl, bobsl, bsldict}, SLT~\cite{gsl, keti, phoenix14t, bobsl}, and SL production~\cite{isl}. These datasets cover a wide range of different sign languages, such as ASL~\cite{asllvd, msasl, wlasl}, BSL~\cite{bsldict, bobsl}, DGS~\cite{phoenix14, phoenix14t}, CSL~\cite{csl100, csl-daily}, GSL~\cite{gsl}, TSL~\cite{bosphorus-sign-22k, autsl}, and ISL~\cite{isl}. Although there are some HKSL resources for educational purposes, there are very few HKSL resources for the development of automatic SLR/SLT models, with the exception of the relatively small dataset mentioned in SignBERT~\cite{zhou2021signbert} and HKU-Portable~\cite{hku.portable.hksl.2022}. 
Both works collected HKSL datasets composed of 50 unique sentences. SignBERT collected data using vision-based technologies, while HKU-Portable employed the Inertial Measurement Unit (IMU) data from smart watches. These datasets, despite their important contributions, are limited in their small amount of unique sentences and restricted vocabulary.
In contrast, our TVB-HKSL-News dataset was collected from TV programs that comprises more diverse sign sentences with a much larger vocabulary size for both SLR and SLT research. 

Several datasets such as \cite{asllvd,msasl,wlasl,bsl-1k,csl500,autsl} have been developed specifically for isolated SLR. While these datasets are beneficial for training models to recognize individual signs, they do not capture the complex co-articulations between signs that are inherent in continuous signing. 
Continuous sign language recognition and translation, which recognize the natural interconnection between signs at the sentence level, provide the opportunity to gain a more comprehensive understanding of sign language expression.
There has been a recent focus on developing datasets for continuous SLR and SLT~\cite{keti, csl100, csl-daily, phoenix14, phoenix14t, bobsl} as shown in~\Cref{table:related}.
Datasets, such as KETI~\cite{keti}, CSL100~\cite{csl100}, and CSL-Daily~\cite{csl-daily}, were created by recording a fixed set of sign sentences in a laboratory setting. Given that the signers perform sentences verbatim in these datasets, the resulting sign language can appear less natural compared to everyday, free-form signing. 
The datasets BOBSL~\cite{bobsl}, PHOENIX-14~\cite{phoenix14}, LSA-T~\cite{lsat}, and PHOENIX-14T~\cite{phoenix14t}, which were sourced from more spontaneous environments like TV news broadcasts, feature a more natural form of signing. However, this way of data collection requires a post-glossing step to obtain the sign labels for SLR, which can be time-consuming. Some previous works~\cite{lsat} lack sign glosses entirely, while others~\cite{bobsl} use an automatic sign gloss spotting method that may introduce errors.
\begin{figure}[t]
	\centering
	\includegraphics[width=\linewidth]{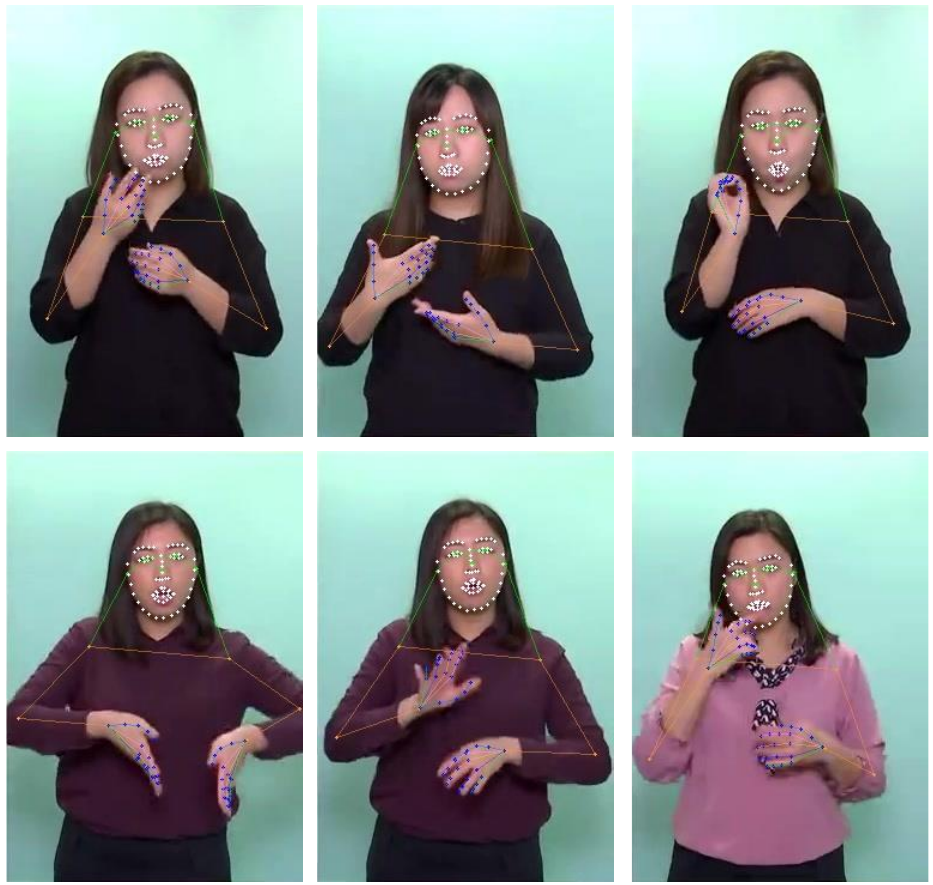}
	\caption{Keypoints extracted from Signer-1/Signer-2 in the first/second row, respectively.}
	\label{fig:kps}
\end{figure}

Our approach to data collection parallels that of BOBSL~\cite{bobsl}, with both datasets being sourced from TV programs via automated or semi-automated pipelines. However, in contrast to BOBSL, our methodology does not leverage the audio track and instead primarily employs computer vision techniques throughout the pipeline.
Although much smaller than BOBSL, our dataset has been carefully and professionally annotated at the gloss level, whereas the gloss annotations in BOBSL are generated through automatic spotting.
Among the SLs listed in Table~\ref{table:related}, our dataset has the largest gloss vocabulary, and is the second largest in the size of SLT vocabulary in terms of words.

%% file: sections/stats.tex
\begin{figure}
	\centering
	\includegraphics[width=\linewidth, trim={50 0 10 0}, clip]{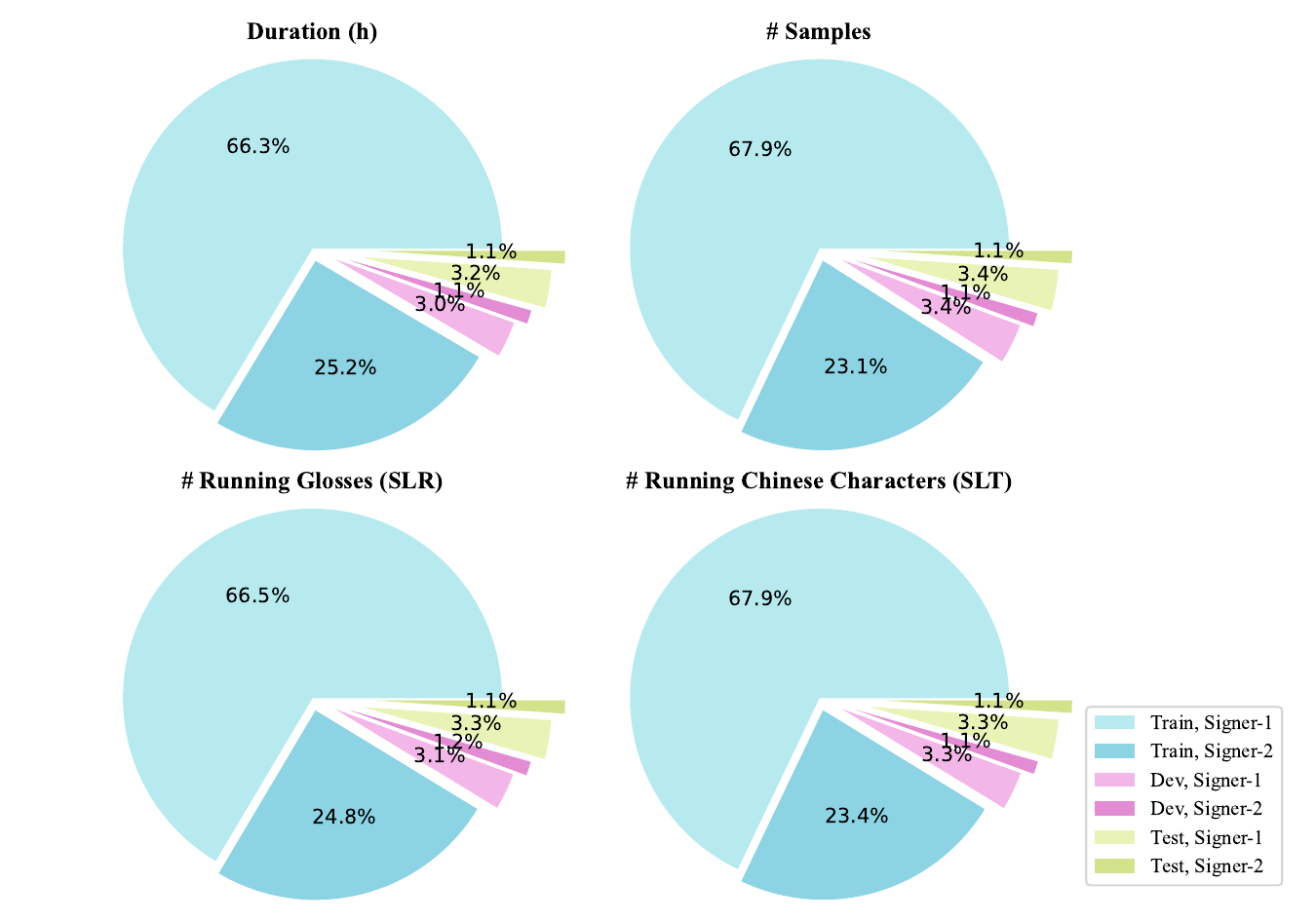}
	\caption{Pie charts showing statistics: number of samples, duration (hours), and running glosses/characters for SLR/SLT tasks across train/dev/test sets.}
	\label{fig:hksl-pie-chart}
\end{figure}

\begin{figure*}[t]
	\centering
	\includegraphics[width=\linewidth, trim={0 10 0 10},clip]{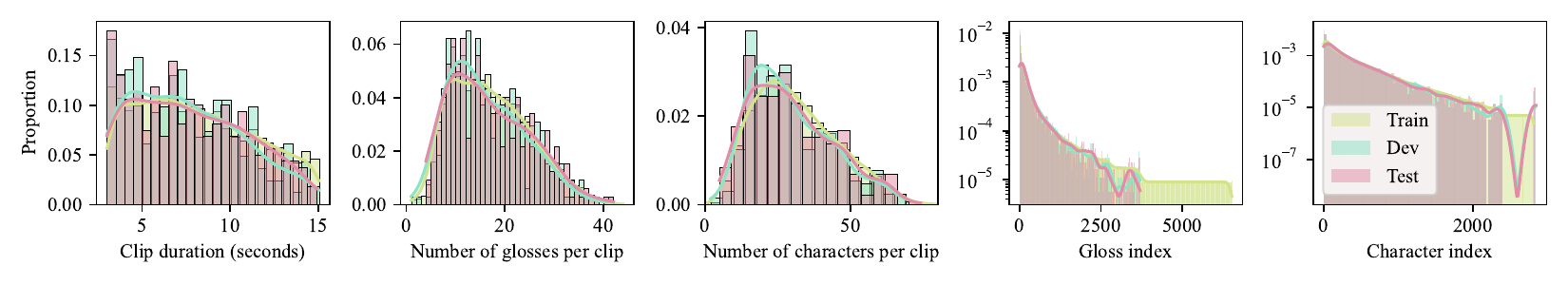}
	\caption{The first three figures show the distribution of clip lengths, number of glosses per clip for SLR, and characters per clip for SLT. The last two figures show the occurrences of particular sign glosses and characters among the samples, sorted by their overall occurrences. Different colors represent different dataset splits.}
	\label{fig:hksl.hist.chart}
\end{figure*}

\section{TVB-HKSL-News Dataset}

The proposed TVB-HKSL-News dataset was sourced from the \textit{TVB News Report with Sign Language program}\footnote{The link to the TVB News Report with Sign Language:~\url{https://news.tvb.com/tc/programme/newsreportwithsignlanguage}.}.
The program covers a wide range of topics, including but not limited to politics, economy, sports, and weather, with each episode running approximately 20 minutes.
Broadcast at a frame rate of 25 fps and a resolution of 1024$\times$576, it features Chinese subtitles and an HKSL interpreter performing sign language interpretation in a fixed 248$\times$360 area on the right side of the screen, alongside the news anchor (see \Cref{fig:hksl-raw}).
%
To build our dataset, we acquired 206 episodes of the program, aired from January 16 to August 17, 2020, with the permission of Television Broadcasts Limited (TVB). These episodes featured simultaneous sign interpretation by two professional HKSL interpreters.

\subsection{Dataset Specifications}

Our dataset, compiled from the source data above, comprises the following four data types:

\noindent \textbf{Sign Video Clips.}~The sign video clips are RGB video segments extracted from the designated region of the sign interpreter on the screen. It has a resolution of 248$\times$360 and a frame rate of 25 fps, with the signer centered in the frame.

\noindent \textbf{Sign Keypoints.}~In accordance with recent research trends that emphasize human body keypoints modeling \cite{stmc, zuo2022c2slr, chentwo, sam-slr, hu2021signbert, zuo2023natural}, we also provide estimated keypoints for each video frame. These keypoints, comprising 68 for the face, 42 for the hands, and 11 for the upper body (as depicted in Figure \ref{fig:kps}), are designed to be robust to variations in signer appearance, hand positions, and shapes.

\noindent \textbf{Chinese Subtitles for SLT.}~We extract textual subtitles from the sign clips using Optical Character Recognition (OCR). These subtitles transcribe the content spoken by the news anchors and are presented in the form of standard spoken Chinese sentences, serving as a resource for SLT research.

\noindent \textbf{Chinese Glosses for SLR.}~For SLR studies, we provide Chinese glosses, obtained from professional sign annotators, for each of the sign clips. These glosses are words that describe the signs verbatim and are sequenced in the order of HKSL, making them suitable for SLR tasks.

\subsection{Dataset Statistics}

The overall statistics of our proposed dataset are presented in \Cref{table:hksl-stats}.
The dataset consists of \numglosses{} glosses for SLR and \numchars{} Chinese characters for SLT.
We divided the dataset into training, development (dev), and test sets according to a rough ratio of 90:5:5. Out-of-vocabulary (OOV) glosses are intentionally avoided in the dev and test sets by rejection sampling. All data splits cover both interpreters, labeled as \textit{Signer-1} and \textit{Signer-2}.
\Cref{fig:hksl-pie-chart} shows the detailed data distribution among the three dataset splits: Signer-1 contributes \numhourssa{} hours of videos, accounting for about 73\% of all sign videos, while Signer-2 contributes \numhourssb{} hours of videos.
The dataset exhibits a wide range of video durations, glosses per segment, and Chinese characters per segment. As shown in \Cref{fig:hksl.hist.chart}, the duration of the video segments ranges from 3--15 seconds, with a mode of 7.88 seconds; the number of glosses per segment ranges from 1--44, with a mode of 16 glosses; the number of Chinese characters per segment ranges from 1--120, with a mode of 29 characters.

%% file: sections/method.tex
\section{Dataset Collection Methodology}

In this section, we outline our data collection methodology for both SLT and SLR. We begin by describing the SLT collection pipeline. Afterwards, we detail the glossing method we adopted to obtain annotations for the SLR task.

\subsection{Data Collection for SLT}

\label{sec:collect}

\noindent \textbf{Obtaining Sign Video Clips.}~
Since the sign language interpreter does not continuously sign throughout the TV news program, we need to perform signer activity detection. To achieve this, we train a video clip binary classifier for each sign language interpreter to detect his/her active and inactive periods.
The classifier is built based on a ResNet18+TCN~\cite{resnet,tcn} model. It takes five video frames as input and produces the probability that the signer is actively signing in the center (i.e., the third) frame.
To train the classifier, we use the data from 60 episodes with one minute per episode, which are manually annotated with activity labels (i.e., 0 and 1).
We use the trained classifier to generate these labels for the remaining videos. The labels help us group consecutive active frames together to create active sign clips. Finally, we filter and retain only clips with durations ranging from 3 to 15 seconds.

\begin{figure}[t]
	\centering
	\includegraphics[width=\linewidth, trim={70 160 90 160},clip]{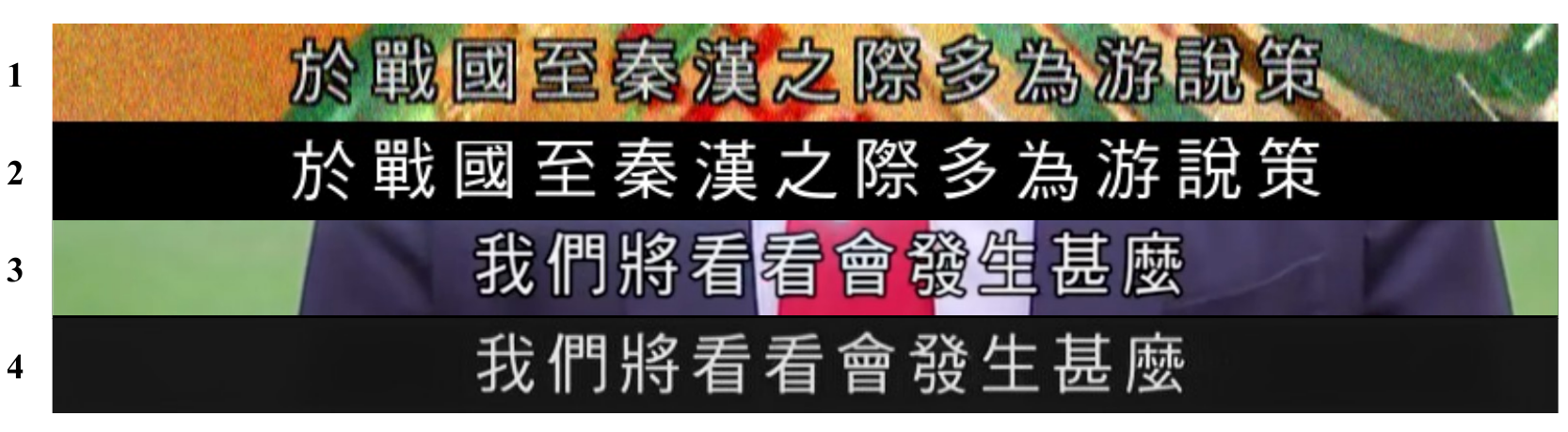}
	\caption{Example of subtitle background removal. Lines 1 and 2 show an example of sample input and target image used for training the U-Net model. The trained model processes Line 3 (input) to produce Line 4 (output) with background removed.}
	\label{fig:background-removal}
\end{figure}

\noindent \textbf{Obtaining Text from Subtitles.}~
Text transcriptions for SLT are obtained from the subtitles that are embedded in a fixed 768$\times$48 region at the bottom of the TV program screen. After cropping this fixed region from the video, we apply the following procedure to extract the text:
\begin{enumerate}
	\item \textit{Subtitle Background Removal.} We first train a background removal model based on the U-Net~\cite{u-net}. The model clears the background of the subtitle with black, as shown in \Cref{fig:background-removal}. To train this model, we first create a synthetic dataset consisting of pairs of source and target images. A source image comprises a non-textual TV program background overlaid with random text, augmented with white noise and Gaussian blur. A target image is a pure black background overlaid with the same text at the same position. The non-textual TV program backgrounds are obtained by randomly cropping a region above the subtitle on the screen, and random text is generated from the zhwiki corpus\footnote{The zhwiki corpus: \url{https://dumps.wikimedia.org/zhwiki/}}.
	      The well-trained model is applied to each frame of the episode specifically within the subtitle region. This process effectively removes the background, yielding clean subtitle videos.
	\item \textit{Subtitle Segmentation.} Subtitle video clips are extracted by identifying rapid transitions in cleaned subtitle videos. These rapid transitions include the appearance, disappearance, and switching of subtitles. To identify such rapid transitions, we apply a temporal Laplacian filter to the intensity values of each pixel across consecutive cleaned frames. We then take the average of all the filtered pixel values in each frame and use it as a measure of the subtitle transition. Peaks in this measure signify rapid transitions in subtitles. We select frames with large transitions via thresholding and use them to segment the video into shorter subtitle clips. Ideally, each of these clips should contain a unique subtitle (including the empty subtitle). Within each subtitle clip, we average all frames. Clips that yield a pure black image, indicating that the subtitle is absent, are subsequently removed.
	\item \textit{Optical character recognition (OCR).} After obtaining the averaged frame of each subtitle video clip, we use the Google OCR API\footnote{Google OCR API: \url{https://cloud.google.com/vision/docs/ocr}.} to recognize the text in the averaged frame.
	\item \textit{Text Re-grouping.}  In practice, we find that the above simple thresholding method can result in oversegmentation, where the same subtitle appears in multiple consecutive video clips. To address this, we iteratively merge adjacent subtitles as long as their edit distance~\cite{levenshtein} is below a threshold. This process can form groups consisting of multiple subtitles. Within each group, we choose the most representative subtitle by selecting the one that has the minimum average edit distance to all the other subtitles in the group. In the end, we obtain the subtitle text along with the precise starting and ending timestamps by considering the frame index of the first and last subtitle frames in the group, respectively.
\end{enumerate}

\noindent \textbf{Aligning Sign and Subtitle Clips.}~
To align the sign clips and subtitle clips, we adopt a Dynamic Time Warping (DTW)-based algorithm. This algorithm treats all the sign clips and subtitle clips within one episode as two time series, and attempts to align them based on the distance between their temporal midpoints. Notably, in our algorithm, we allow multiple subtitle clips to align with a single sign clip, as subtitle clips are typically shorter than sign clips.

\noindent \textbf{Keypoint Extraction from Sign Videos.}~
To extract the facial, hand and body keypoints from sign videos, we use HRNet~\cite{hrnet}, which is pre-trained on COCO-WholeBody~\cite{coco-wb}, following the approach in \cite{chentwo,hu2021signbert,zuo2023natural,zuo2024towards}. The model predicts 133 keypoints, from which we discard 6 lower body keypoints and 6 foot keypoints, as they are irrelevant to sign language tasks and our sign frames only show the upper body of the signers. This process leads to a final count of 121 keypoints per frame, including those for the upper body, hands, and face.

\subsection{Glossing for SLR}

Sign languages have their own rules for word order, sentence structure, and grammar that are different from those of spoken languages. Therefore, subtitles for spoken languages cannot be directly used as glosses for sign language recognition, where monotonic gloss sequences are required as training targets.
To obtain accurate gloss annotations of the signed clips collected, we hired HKSL signers from SLCO Community Resources\footnote{SLCO Community Resources: \url{https://www.slco.org.hk}.} to annotate them.
The annotators are fluent in both HKSL and Cantonese (Chinese).
To facilitate the annotation process, we have developed a software tool for the annotators. The software presents sign language video clips side by side with the corresponding subtitle text, which is derived from the previous pipeline. Annotators can efficiently gloss the sign language videos using the tool with the subtitle text as a reference.
During the annotation process, the annotator glosses the signs in the video verbatim, from left to right.
In addition to the common signs that can be described using single Chinese words, there are signs that require special attention.
The glossing of these special cases is dealt with using the following rules:
\begin{enumerate}
	\item \textit{Compound signs} are annotated in the form of \textit{X+Y}. They are formed by combining two or more signs to create a phrase of distinct meaning. During post-processing, we split these signs into \textit{X Y}.
	\item \textit{Ill-performed signs} are signs that are not signed correctly but can still be recognized by the annotators. They are annotated as \textit{X(?)}. During post-processing, we assume they are the truly \textit{X}.
	\item \textit{Many-to-one glossing} occurs when multiple signs are glossed with the same term, represented as \textit{X(1)}, \textit{X(2)}, and so on. We choose the common glossing, \textit{X}, for all these variations.
	\item \textit{One-to-many glossing} arises when a single sign can be transcribed into multiple glosses due to the availability of various suitable gloss words. This is annotated as a homosign group: \textit{X(=Y=Z=...)}. During training, we select the gloss with the highest compound count for each group (e.g., A+B+C has the compound count of three), or the lowest lexicographical order if there is a tie. In the testing phase, we proceed by globally merging all homosign groups that share at least one common element. The selection of the representative gloss for the newly merged group follows the same strategy used during training. Before computing the score, we replace glosses appearing in any homosign group with its representative gloss in both the hypothesis and reference sentences for each sample.
\end{enumerate}
These post-processing rules simplify raw annotations and result in cleaner and more consistent glossing of signs, while preserving the necessary information for the SLR task\footnote{The original raw annotations are available so that users may investigate their own post-processing rules.}.

%% file: sections/expr_v5.8.tex
\begin{table*}[t]
	\caption{Baseline results for SLR.}
	\label{table:slr}
	\centering
	\resizebox{0.6\linewidth}{!}{
		\begin{tabular}{l|cc|cc}
			\toprule
			\multirow{2}{*}{Method}             & \multicolumn{2}{c|}{Modality} & \multicolumn{2}{c}{WER (\%)}                             \\
			                                    & Video                         & Keypoints                    & Dev         & Test        \\
			\midrule
			S3D \cite{s3d, chentwo}             &                               & \cmark                       & 45.73       & 44.56       \\
			S3D \cite{s3d, chentwo}             & \cmark                        &                              & 39.59       & 38.63       \\
			VLT \cite{zuo2022c2slr}             & \cmark                        &                              & 35.89       & 36.18       \\
			$\text{C}^2$SLR \cite{zuo2022c2slr} & \cmark                        & \cmark                       & 35.43       & 35.78       \\
			TwoStream-SLR \cite{chentwo}        & \cmark                        & \cmark                       & {\bf 34.52} & {\bf 34.08} \\
			\bottomrule
		\end{tabular}
	}
\end{table*}

\begin{table*}[t]
	\caption{Baseline results for SLT. (R: ROUGE, B: BLEU.)}
	\label{table:slt}
	\centering
	\resizebox{\linewidth}{!}{
		\begin{tabular}{l|ccccc|ccccc}
			\toprule
			\multirow{2}{*}{Method} & \multicolumn{5}{c|}{Dev} & \multicolumn{5}{c}{Test}                                                                                \\
			                        & R                        & B-1                      & B-2            & B-3            & B-4            & R & B-1 & B-2 & B-3 & B-4 \\
			\midrule
			S3D (video) \cite{s3d, chentwo}
			                        & 18.64                    & 21.98                    & 15.17          & 11.18          & 8.79
			                        & 21.61                    & 25.39                    & 18.59          & 14.57          & 12.10                                      \\
			S3D (keypoints) \cite{s3d, chentwo}
			                        & 15.65                    & 18.18                    & 11.58          & 8.09           & 6.22
			                        & 16.42                    & 19.93                    & 13.72          & 10.41          & 8.48                                       \\
			TwoStream-SLT \cite{chentwo}
			                        & \textbf{38.12}           & \textbf{43.22}           & \textbf{33.44} & \textbf{26.04} & \textbf{21.00}
			                        & \textbf{39.80}           & \textbf{44.68}           & \textbf{35.27} & \textbf{28.29} & \textbf{23.58}                             \\
			\bottomrule
		\end{tabular}
	}
\end{table*}

\section{Experiments}

Here, we present our experiments on SLR and SLT. We first introduce the baseline models and the evaluation metrics that are used to assess their performance. We then report our experimental results on the full dataset and provide a qualitative analysis of the outcomes. Finally, we evaluate the impact of the amount of training data under a signer-dependent setting for a single signer.
We hope that the experimental results will serve as benchmarks for future research on this dataset.

\subsection{Baselines and Evaluation Metrics}
\noindent \textbf{Video-based Baselines.}
Most existing deep-learning-based SLR models adopt 2D- or 3D-CNNs \cite{vggnet, resnet, I3D, s3d} for visual feature extraction.
A sequential module, which is composed of, for example, temporal CNNs \cite{tcn}, Transformer \cite{transformer}, or a mixture of them \cite{qanet}, is optionally appended to the CNNs for temporal modeling.
In this paper, we choose S3D \cite{s3d} for the implementation of 3D-CNNs, which is adopted as the backbone network of the video/keypoint encoder in TwoStream-SLR \cite{chentwo} (the current best model on several widely adopted SLR benchmarks \cite{phoenix14, phoenix14t, csl-daily}) due to its strong spatial-temporal modeling capability and high efficiency.
Besides, we also implement VLT \cite{zuo2022c2slr} as our sequential module with a mixed architecture, which is the backbone network of $\text{C}^2$SLR \cite{zuo2022c2slr}; it uses VGG11 \cite{vggnet} to extract frame-wise visual features and a local transformer (LT) \cite{zuo2022local, qanet} to model temporal dependencies.

\noindent \textbf{Baselines that Utilize Keypoints.}
Substantial visual redundancy in RGB videos may lead video-based SLR models to overlook the key information for SL understanding \cite{chentwo}.
Besides, the performance of video-based models may also suffer from large variations in video backgrounds and signer appearances.
To obtain better representations, some SLR works \cite{chentwo, zuo2022c2slr, stmc, hu2021signbert, sam-slr} also make use of keypoints as complementary information in their models.
In this paper, we first implement $\text{C}^2$SLR \cite{zuo2022c2slr} which utilizes keypoint heatmaps of signers' faces and hands as a guidance for a spatial attention module to enforce the visual module to focus on informative regions of a signing frame.
Second, we implement TwoStream-SLR \cite{chentwo}, which also represents keypoints as a sequence of heatmaps and shares an identical architecture (S3D) between the video and keypoint streams.
Finally, TwoStream-SLT \cite{chentwo}, which simply appends an MLP and a translation network to each head of TwoStream-SLR is used for SLT.\footnote{We only focus on the Sign2Text setting, which directly translates sign videos into spoken languages. It has also been proven that it outperforms the two-stage Sign2Gloss2Text setting \cite{chen2022simple, chentwo}.}
We implement all baselines by following their original works. All models are trained on 8 NVIDIA Tesla V100 GPUs with a batch size of 8.

\noindent \textbf{Evaluation Metrics.}
Following the usual practice~\cite{chentwo, zuo2022c2slr, stmc, phoenix14t, phoenix14}, we adopt word (gloss) error rate (WER) for SLR, and BLEU \cite{papineni2002bleu} and ROUGE-L \cite{rouge} for SLT. Lower WER means better SLR performance while higher BLEU and ROUGE-L indicate better SLT performance.

\begin{CJK*}{UTF8}{bsmi}
	\begin{table*}[ht]
		\caption{Qualitative results on TVB-HKSL-News. For SLR, we use different colors to represent \sub{substitutions}, \del{deletions}, and \ins{insertions}, respectively.}
		\label{tab:qual}
		\centering
		\resizebox{0.92\linewidth}{!}{
			\begin{tabular}{l|l|c}
				\toprule
				\textbf{SLR}                           &                                                                                                                                          & WER (\%)               \\
				\midrule
				\multirow{2}{*}{Ground Truth}          & 昨天\ 溫度\ 二\ 十\ 有\ 濕\ 百分比\ 七\ 六                                                                                               & \multirow{2}{*}{-}     \\
				                                       & (Yesterday Temperature Two Ten Has Humidity Percentage Seven Six)                                                                        &                        \\
				\midrule
				\multirow{2}{*}{Pred. (S3D Video)}     & \sub{以前}\ 溫度\ \sub{小}\ \del{*}\ 有\ 濕\ 百分比\ 七\ 六                                                                              & \multirow{2}{*}{33.33} \\
				                                       & (\sub{Previously} Temperature \sub{Low} \del{*} Humidity Percentage Seven Six)                                                           &                        \\
				\midrule
				\multirow{2}{*}{Pred. (S3D Keypoint)}  & \del{**}\ 溫度\ \del{*}\ 十\ \del{*}\ 濕\ 百分比\ 七\ \ins{九}\ 六                                                                       & \multirow{2}{*}{44.44} \\
				                                       & (\del{**} Temperature \del{*} Ten \del{*} Humidity Percentage Seven \ins{Nine} Six)                                                      &                        \\
				\midrule
				\multirow{2}{*}{Pred. (TwoStream-SLR)} & \del{**}\ 溫度\ 二\ 十\ \del{*}\ 濕\ 百分比\ 七\ 六                                                                                      & \multirow{2}{*}{22.22} \\
				                                       & (\del{**} Temperature Two Ten \del{*} Humidity Percentage Seven Six)                                                                     &                        \\
				\midrule
				\midrule

				\textbf{SLT}                           &                                                                                                                                          & BLEU-4                 \\
				\midrule
				\multirow{2}{*}{Ground Truth}          & 而輸入個案有一宗是一名印度海員                                                                                                           & \multirow{2}{*}{-}     \\
				                                       & (One of the imported cases was an Indian seafarer.)                                                                                      &                        \\
				\midrule
				\multirow{2}{*}{Pred. (S3D Video)}     & \underline{至於}輸入個案有\underline{11宗}\underline{包括}印度海員                                                                       & \multirow{2}{*}{39.38} \\
				                                       & (\underline{Regarding} the imported cases, there are \underline{eleven} cases \underline{including} an Indian seafarer.)                 &                        \\
				\midrule
				\multirow{2}{*}{Pred. (S3D Keypoint)}  & 有\sout{輸入個案}滯留印度的\underline{港人}\sout{海員}知道\underline{賭場重開}                                                           & \multirow{2}{*}{0.00}  \\
				                                       & (Some \sout{imported cases} \underline{Hong Kong people} \sout{seafarer} stranded in India know that \underline{casinos are reopening}.) &                        \\
				\midrule
				\multirow{2}{*}{Pred. (TwoStream-SLT)} & \underline{至於}輸入個案有一名印度海員                                                                                                   & \multirow{2}{*}{63.86} \\
				                                       & (\underline{Regarding} the imported cases, there is one Indian seafarer.)                                                                &                        \\

				\bottomrule
			\end{tabular}}
	\end{table*}
\end{CJK*}

\subsection{SLR and SLT Baseline Results}
\noindent \textbf{SLR.}
The baseline results for SLR are shown in Table \ref{table:slr}.
Using S3D with keypoints as the baseline, video inputs can lead to much better performance
than keypoint heatmap inputs (44.56\% $\rightarrow$ 38.63\% on the test set).
The results are reasonable since RGB videos contain much more information than sparse keypoints.
For video-based baselines, VLT outperforms S3D by 2.45\% on the test set. The result suggests that a sequential module, e.g., a local transformer, with both local and global context modeling capability is preferred.
For baselines that utilize keypoints,  $\text{C}^2$SLR uses pre-extracted keypoint heatmaps to enforce the spatial attention module to focus on informative sign regions.
The mechanism is also effective on the proposed HKSL dataset with an improvement of 0.46\% on the dev set over the best video-based baseline.
The lowest WERs (34.52\%/34.08\% on the dev/test set) come from the TwoStream-SLR model, which better exploits the benefits of keypoints by developing a dual-S3D architecture that takes both videos and keypoint heatmaps as inputs.
It significantly improves the performance of the S3D baseline by 5.07\%/4.55\% on the dev/test set, respectively.

\noindent \textbf{SLT.}
Table \ref{table:slt} provides several SLT baseline results.
The TwoStream-SLT model achieves a BLEU-4 score of 23.58 on the test set. It outperforms the single-stream models by a large margin, and thus validates the effectiveness of the joint modeling of both videos and keypoints.
The SLT performance on our new dataset is comparable to that on CSL-Daily \cite{chen2022simple} (25.79 on the test set). %
The written texts in both datasets are Chinese, but the character vocabulary size of our dataset is about 20\% larger, and our word vocabulary size is more than double of CSL-Daily's, and yet there are only 2 signers in our training set whereas there are 10 signers in CSL-Daily.

\noindent \textbf{Qualitative Results.}
SLR and SLT results on some test data using the strongest baseline, TwoStream-SLR/SLT, are shown in Table \ref{tab:qual}.
It is clear that the two-stream model can yield more accurate gloss/text predictions than both the single-stream models, which only take RGB videos or keypoint heatmaps as inputs.
The results also suggest that the two modalities can complement each other.

\subsection{Evaluation on Single-signer SLR/SLT}

\begin{figure}[t]
	\centering
	\includegraphics[width=0.95\linewidth]{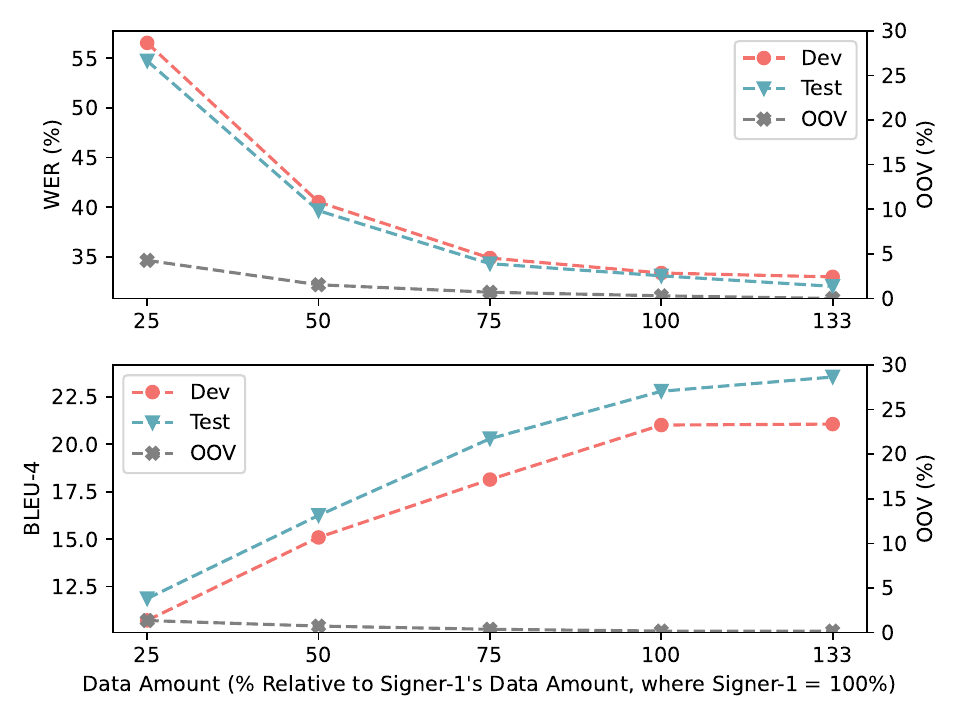}
	\caption{SLR and SLT performance of TwoStream-SLR/SLT with different amount of Signer-1's training data. 133\% means that we use all the training data of both Signer-1 and Signer-2 (whose amount of data is about 33\% of Signer-1's) but evaluate the trained model only on Signer-1's dev/test data.}
	\label{fig:single}
\end{figure}

One key characteristic of our HKSL dataset is that it has a large amount of data for a single-signer, Signer-1, the major signer in the dataset with \numhourssa{} hours of sign videos that consists of \numsamplessa{} samples. This enables studies on how well SLR/SLT performance can be achieved under the signer-dependent setting.
We study the effect of the amount of training data on SLR/SLT performance using the TwoStream-SLR/SLT model and only Signer-1's training data
(with the exception of the ``133\% experiment''). Performance is evaluated only on Signer-1's test data.
The results are shown in Figure \ref{fig:single} together with the corresponding OOV ratios (grey lines). Notice that since this evaluation uses Signer-1's data only, the results shown in Figure \ref{fig:single} are different from results shown in Table \ref{table:slr} and \ref{table:slt} (as the latter use data from all signers).

First, we notice that the maximum OOV ratio is just 4.27\% and 1.35\% for SLR and SLT, respectively, when using only 25\% training data.
Thus, the effect of OOV is negligible on the SLR/SLT performance.
Second, we find that, as expected, the model performance on both SLR and SLT improves when more training data are available.
Third, the addition of more training data from another signer (shown as 133\% training data in Figure \ref{fig:single}) can still improve the performance.
We believe the use of keypoint information in the TwoStream-SLR/SLT model makes the SLR/SLT performance more robust to signer appearances.
Finally, SLR performance converges with the use of 75\% of the training data (which is around 8 hours of sign videos) but SLT performance improves linearly with the amount of training data.
For instance, increasing the amount of Signer-1's data from 75\% to 100\% can only lead to a WER reduction of 1.22\% (34.30\% $\rightarrow$ 33.08\%, or 3.56\% relative) on the test set. On the other hand, the corresponding BLEU-4 improvement is 2.50 (20.29 $\rightarrow$ 22.79, or 12.32\% relative).
The results suggest that around 8 hours of training data may be sufficient for signer-dependent SLR modeling in our setting, but SLT still benefits from much more data than SLR. This can serve as a guidance for future research on signer-dependent SLR/SLT.

%% file: sections/conclusion.tex
\section{Conclusion}
This work introduces the TVB-HKSL-News dataset, which is a valuable resource for advancing research in continuous SLR and SLT. The dataset also provides a large amount of data for one of the signers with manual gloss annotations, enabling researchers to investigate new modeling methods and evaluate the impact of the amount of training data needed for the SLR and SLT of a single signer.
Furthermore, the automated data collection pipeline introduced in the paper makes it easier to scale up the collection of a large amount of data for SLT in the future if sign-interpreted videos with subtitles are available.
Moreover, the presented results of some state-of-the-art SLR/SLT models on the dataset serve as the benchmarks for future research on the dataset.


\section{Ethical Consideration}

During the creation and the use of the TVB-HKSL-News sign language dataset in our project, we have obtained informed consent from the sign language interpreters involved in the dataset. They are aware of their data being used for academic research purposes.
Additionally, we have implemented a procedure to make the dataset accessible for academic research. The interested parties are required to sign an agreement with Television Broadcasts Limited (TVB), ensuring that the dataset will be used only for academic research.

\section{Dataset Availability Statement}

The complete dataset will be hosted on a dedicated server.
Authorized users can download it using a specific URL and password, which can be requested at \url{https://tvb-hksl-news.github.io/}.
To gain authorization, users must first sign an agreement that will be available on our dataset website, ensuring the dataset's appropriate use.
The copyrights for all materials in the TVB-HKSL-News dataset belongs to Television Broadcasts Limited (TVB). We have obtained permission from TVB for its distribution strictly for {\bf academic research} and the data {\bf should not be used for any other commercial or non-commercial purposes}.

%% file: main.bbl
\begin{thebibliography}{48}
\expandafter\ifx\csname natexlab\endcsname\relax\def\natexlab#1{#1}\fi

\bibitem[{Adaloglou et~al.(2022)Adaloglou, Chatzis, Papastratis, Stergioulas, Papadopoulos, Zacharopoulou, Xydopoulos, Atzakas, Papazachariou, and Daras}]{gsl}
Nikolas Adaloglou, Theocharis Chatzis, Ilias Papastratis, Andreas Stergioulas, Georgios~Th. Papadopoulos, Vassia Zacharopoulou, George~J. Xydopoulos, Klimnis Atzakas, Dimitris Papazachariou, and Petros Daras. 2022.
\newblock A comprehensive study on deep learning-based methods for sign language recognition.
\newblock \emph{{IEEE} TMM}, 24:1750--1762.

\bibitem[{Albanie et~al.(2020)Albanie, Varol, Momeni, Afouras, Chung, Fox, and Zisserman}]{bsl-1k}
Samuel Albanie, G{\"{u}}l Varol, Liliane Momeni, Triantafyllos Afouras, Joon~Son Chung, Neil Fox, and Andrew Zisserman. 2020.
\newblock {BSL-1K:} scaling up co-articulated sign language recognition using mouthing cues.
\newblock In \emph{ECCV}, volume 12356, pages 35--53. Springer.

\bibitem[{Albanie et~al.(2021)Albanie, Varol, Momeni, Bull, Afouras, Chowdhury, Fox, Woll, Cooper, McParland, and Zisserman}]{bobsl}
Samuel Albanie, G{\"{u}}l Varol, Liliane Momeni, Hannah Bull, Triantafyllos Afouras, Himel Chowdhury, Neil Fox, Bencie Woll, Rob Cooper, Andrew McParland, and Andrew Zisserman. 2021.
\newblock {BBC-Oxford British} sign language dataset.
\newblock \emph{CoRR}, abs/2111.03635.

\bibitem[{Athitsos et~al.(2008)Athitsos, Neidle, Sclaroff, Nash, Stefan, Yuan, and Thangali}]{asllvd}
Vassilis Athitsos, Carol Neidle, Stan Sclaroff, Joan~P. Nash, Alexandra Stefan, Quan Yuan, and Ashwin Thangali. 2008.
\newblock The {A}merican sign language lexicon video dataset.
\newblock In \emph{CVPR}, pages 1--8.

\bibitem[{Bianco et~al.(2022)Bianco, R{\'{\i}}os, Ronchetti, Quiroga, Stanchi, Hasperu{\'{e}}, and Rosete}]{lsat}
Pedro~Dal Bianco, Gast{\'{o}}n R{\'{\i}}os, Franco Ronchetti, Facundo~Manuel Quiroga, Oscar Stanchi, Waldo Hasperu{\'{e}}, and Alejandro Rosete. 2022.
\newblock {LSA-T:} the first continuous argentinian sign language dataset for sign language translation.
\newblock In \emph{Advances in Artificial Intelligence - {IBERAMIA} 2022 - 17th Ibero-American Conference on AI, Cartagena de Indias, Colombia, November 23-25, 2022, Proceedings}, volume 13788 of \emph{Lecture Notes in Computer Science}, pages 293--304. Springer.

\bibitem[{Camg{\"{o}}z et~al.(2018)Camg{\"{o}}z, Hadfield, Koller, Ney, and Bowden}]{phoenix14t}
Necati~Cihan Camg{\"{o}}z, Simon Hadfield, Oscar Koller, Hermann Ney, and Richard Bowden. 2018.
\newblock Neural sign language translation.
\newblock In \emph{CVPR}, pages 7784--7793.

\bibitem[{Carreira and Zisserman(2017)}]{I3D}
Joao Carreira and Andrew Zisserman. 2017.
\newblock Quo vadis, action recognition? a new model and the kinetics dataset.
\newblock In \emph{CVPR}.

\bibitem[{Chai et~al.(2014)Chai, Wang, and Chen}]{devisign}
Xiujuan Chai, Hanjie Wang, and Xilin Chen. 2014.
\newblock The devisign large vocabulary of {Chinese} sign language database and baseline evaluations.
\newblock In \emph{Technical report VIPL-TR-14-SLR-001. Key Lab of Intelligent Information Processing of Chinese Academy of Sciences (CAS)}. Institute of Computing Technology.

\bibitem[{Chen et~al.(2022{\natexlab{a}})Chen, Wei, Sun, Wu, and Lin}]{chen2022simple}
Yutong Chen, Fangyun Wei, Xiao Sun, Zhirong Wu, and Stephen Lin. 2022{\natexlab{a}}.
\newblock A simple multi-modality transfer learning baseline for sign language translation.
\newblock In \emph{CVPR}, pages 5120--5130.

\bibitem[{Chen et~al.(2022{\natexlab{b}})Chen, Zuo, Wei, Wu, Liu, and Mak}]{chentwo}
Yutong Chen, Ronglai Zuo, Fangyun Wei, Yu~Wu, Shujie Liu, and Brian Mak. 2022{\natexlab{b}}.
\newblock Two-stream network for sign language recognition and translation.
\newblock In \emph{NeurIPS}.

\bibitem[{He et~al.(2016)He, Zhang, Ren, and Sun}]{resnet}
Kaiming He, Xiangyu Zhang, Shaoqing Ren, and Jian Sun. 2016.
\newblock Deep residual learning for image recognition.
\newblock In \emph{CVPR}, pages 770--778.

\bibitem[{Hu et~al.(2021)Hu, Zhao, Zhou, Wang, and Li}]{hu2021signbert}
Hezhen Hu, Weichao Zhao, Wengang Zhou, Yuechen Wang, and Houqiang Li. 2021.
\newblock {SignBERT}: Pre-training of hand-model-aware representation for sign language recognition.
\newblock In \emph{ICCV}, pages 11087--11096.

\bibitem[{Huang et~al.(2019)Huang, Zhou, Li, and Li}]{csl500}
Jie Huang, Wengang Zhou, Houqiang Li, and Weiping Li. 2019.
\newblock Attention-based {3D-CNNs} for large-vocabulary sign language recognition.
\newblock \emph{{IEEE} TCSVT.}, 29(9):2822--2832.

\bibitem[{Huang et~al.(2018)Huang, Zhou, Zhang, Li, and Li}]{csl100}
Jie Huang, Wengang Zhou, Qilin Zhang, Houqiang Li, and Weiping Li. 2018.
\newblock Video-based sign language recognition without temporal segmentation.
\newblock In \emph{AAAI}, pages 2257--2264.

\bibitem[{Jiang et~al.(2021)Jiang, Sun, Wang, Bai, Li, and Fu}]{sam-slr}
Songyao Jiang, Bin Sun, Lichen Wang, Yue Bai, Kunpeng Li, and Yun Fu. 2021.
\newblock Skeleton aware multi-modal sign language recognition.
\newblock In \emph{CVPRW}, pages 3413--3423.

\bibitem[{Jiao et~al.(2018)Jiao, Sun, and Sun}]{jiao2018LAC}
Zhenyu Jiao, Shuqi Sun, and Ke~Sun. 2018.
\newblock \href {https://arxiv.org/abs/1807.01882} {{C}hinese lexical analysis with deep bi-gru-crf network}.
\newblock \emph{arXiv preprint arXiv:1807.01882}.

\bibitem[{Jin et~al.(2020)Jin, Xu, Xu, Wang, Liu, Qian, Ouyang, and Luo}]{coco-wb}
Sheng Jin, Lumin Xu, Jin Xu, Can Wang, Wentao Liu, Chen Qian, Wanli Ouyang, and Ping Luo. 2020.
\newblock Whole-body human pose estimation in the wild.
\newblock In \emph{ECCV}, pages 196--214.

\bibitem[{Joze and Koller(2019)}]{msasl}
Hamid Reza~Vaezi Joze and Oscar Koller. 2019.
\newblock {MS-ASL:} {A} large-scale data set and benchmark for understanding american sign language.
\newblock In \emph{BMVC}, page 100.

\bibitem[{Kapoor et~al.(2021)Kapoor, Mukhopadhyay, Hegde, Namboodiri, and Jawahar}]{isl}
Parul Kapoor, Rudrabha Mukhopadhyay, Sindhu~B. Hegde, Vinay~P. Namboodiri, and C.~V. Jawahar. 2021.
\newblock Towards automatic speech to sign language generation.
\newblock In \emph{Interspeech}, pages 3700--3704.

\bibitem[{Ko et~al.(2018)Ko, Kim, Jung, and Cho}]{keti}
Sang{-}Ki Ko, Chang~Jo Kim, Hyedong Jung, and Choong~Sang Cho. 2018.
\newblock \href {http://arxiv.org/abs/1811.11436} {Neural sign language translation based on human keypoint estimation}.
\newblock \emph{CoRR}, abs/1811.11436.

\bibitem[{Koller et~al.(2015)Koller, Forster, and Ney}]{phoenix14}
Oscar Koller, Jens Forster, and Hermann Ney. 2015.
\newblock Continuous sign language recognition: Towards large vocabulary statistical recognition systems handling multiple signers.
\newblock \emph{CVIU}, 141:108--125.

\bibitem[{Lea et~al.(2016)Lea, Vidal, Reiter, and Hager}]{tcn}
Colin Lea, Ren{\'{e}} Vidal, Austin Reiter, and Gregory~D. Hager. 2016.
\newblock Temporal convolutional networks: {A} unified approach to action segmentation.
\newblock In \emph{ECCVW}, Lecture Notes in Computer Science, pages 47--54.

\bibitem[{Levenshtein et~al.(1966)}]{levenshtein}
Vladimir~I Levenshtein et~al. 1966.
\newblock Binary codes capable of correcting deletions, insertions, and reversals.
\newblock In \emph{Soviet physics doklady}, volume~10, pages 707--710. Soviet Union.

\bibitem[{Li et~al.(2020)Li, Opazo, Yu, and Li}]{wlasl}
Dongxu Li, Cristian~Rodriguez Opazo, Xin Yu, and Hongdong Li. 2020.
\newblock Word-level deep sign language recognition from video: {A} new large-scale dataset and methods comparison.
\newblock In \emph{WACV}, pages 1448--1458.

\bibitem[{Li et~al.(2019)Li, Meng, Sun, Han, Yuan, and Li}]{why-slt-character}
Xiaoya Li, Yuxian Meng, Xiaofei Sun, Qinghong Han, Arianna Yuan, and Jiwei Li. 2019.
\newblock Is word segmentation necessary for deep learning of {C}hinese representations?
\newblock In \emph{ACL}, pages 3242--3252.

\bibitem[{Lin(2004)}]{rouge}
Chin-Yew Lin. 2004.
\newblock Rouge: A package for automatic evaluation of summaries.
\newblock In \emph{Text summarization branches out}, pages 74--81.

\bibitem[{Momeni et~al.(2020)Momeni, Varol, Albanie, Afouras, and Zisserman}]{bsldict}
Liliane Momeni, G{\"{u}}l Varol, Samuel Albanie, Triantafyllos Afouras, and Andrew Zisserman. 2020.
\newblock Watch, read and lookup: Learning to spot signs from multiple supervisors.
\newblock In \emph{ACCV}, volume 12627, pages 291--308.

\bibitem[{{\"{O}}zdemir et~al.(2020){\"{O}}zdemir, Kindiroglu, Camg{\"{o}}z, and Akarun}]{bosphorus-sign-22k}
Ogulcan {\"{O}}zdemir, Ahmet~Alp Kindiroglu, Necati~Cihan Camg{\"{o}}z, and Lale Akarun. 2020.
\newblock \href {http://arxiv.org/abs/2004.01283} {{BosphorusSign22k} sign language recognition dataset}.
\newblock \emph{CoRR}, abs/2004.01283.

\bibitem[{Papineni et~al.(2002)Papineni, Roukos, Ward, and Zhu}]{papineni2002bleu}
Kishore Papineni, Salim Roukos, Todd Ward, and Wei-Jing Zhu. 2002.
\newblock {BLEu}: A method for automatic evaluation of machine translation.
\newblock In \emph{ACL}, pages 311--318.

\bibitem[{Ronneberger et~al.(2015)Ronneberger, Fischer, and Brox}]{u-net}
Olaf Ronneberger, Philipp Fischer, and Thomas Brox. 2015.
\newblock U-net: Convolutional networks for biomedical image segmentation.
\newblock In \emph{MICCAI}, volume 9351, pages 234--241.

\bibitem[{Sehyr et~al.(2021)Sehyr, Caselli, Cohen-Goldberg, and Emmorey}]{asl-lex-2.0}
Zed~Sevcikova Sehyr, Naomi Caselli, Ariel~M Cohen-Goldberg, and Karen Emmorey. 2021.
\newblock The {ASL-LEX} 2.0 project: A database of lexical and phonological properties for 2,723 signs in american sign language.
\newblock \emph{The Journal of Deaf Studies and Deaf Education}, 26(2):263--277.

\bibitem[{Simonyan and Zisserman(2015)}]{vggnet}
Karen Simonyan and Andrew Zisserman. 2015.
\newblock Very deep convolutional networks for large-scale image recognition.
\newblock In \emph{ICLR}.

\bibitem[{Sincan and Keles(2020)}]{autsl}
Ozge~Mercanoglu Sincan and Hacer~Yalim Keles. 2020.
\newblock {AUTSL:} {A} large scale multi-modal {T}urkish {S}ign {L}anguage dataset and baseline methods.
\newblock \emph{{IEEE} Access}, 8:181340--181355.

\bibitem[{Sun(2012)}]{jieba}
Junyi Sun. 2012.
\newblock Jieba chinese word segmentation tool.

\bibitem[{Sun et~al.(2019)Sun, Xiao, Liu, and Wang}]{hrnet}
Ke~Sun, Bin Xiao, Dong Liu, and Jingdong Wang. 2019.
\newblock Deep high-resolution representation learning for human pose estimation.
\newblock In \emph{CVPR}, pages 5693--5703.

\bibitem[{{TVB News}(2023)}]{tvbnews}
{TVB News}. 2023.
\newblock News report with sign language.
\newblock \url{https://news.tvb.com/tc/programme/newsreportwithsignlanguage}.

\bibitem[{Vaswani et~al.(2017)Vaswani, Shazeer, Parmar, Uszkoreit, Jones, Gomez, Kaiser, and Polosukhin}]{transformer}
Ashish Vaswani, Noam Shazeer, Niki Parmar, Jakob Uszkoreit, Llion Jones, Aidan~N. Gomez, Lukasz Kaiser, and Illia Polosukhin. 2017.
\newblock Attention is all you need.
\newblock In \emph{NeurIPS}, pages 5998--6008.

\bibitem[{Viitaniemi et~al.(2014)Viitaniemi, Jantunen, Savolainen, Karppa, and Laaksonen}]{s-pot}
Ville Viitaniemi, Tommi Jantunen, Leena Savolainen, Matti Karppa, and Jorma Laaksonen. 2014.
\newblock S-pot - a benchmark in spotting signs within continuous signing.
\newblock In \emph{LREC}, pages 1892--1897.

\bibitem[{Xie et~al.(2018)Xie, Sun, Huang, Tu, and Murphy}]{s3d}
Saining Xie, Chen Sun, Jonathan Huang, Zhuowen Tu, and Kevin Murphy. 2018.
\newblock Rethinking spatiotemporal feature learning: Speed-accuracy trade-offs in video classification.
\newblock In \emph{ECCV}, pages 305--321.

\bibitem[{Yu et~al.(2018)Yu, Dohan, Luong, Zhao, Chen, Norouzi, and Le}]{qanet}
Adams~Wei Yu, David Dohan, Minh{-}Thang Luong, Rui Zhao, Kai Chen, Mohammad Norouzi, and Quoc~V. Le. 2018.
\newblock {QANet}: Combining local convolution with global self-attention for reading comprehension.
\newblock In \emph{ICLR}.

\bibitem[{Zhou et~al.(2021{\natexlab{a}})Zhou, Zhou, Qi, Pu, and Li}]{csl-daily}
Hao Zhou, Wengang Zhou, Weizhen Qi, Junfu Pu, and Houqiang Li. 2021{\natexlab{a}}.
\newblock Improving sign language translation with monolingual data by sign back-translation.
\newblock In \emph{CVPR}, pages 1316--1325.

\bibitem[{Zhou et~al.(2020)Zhou, Zhou, Zhou, and Li}]{stmc}
Hao Zhou, Wengang Zhou, Yun Zhou, and Houqiang Li. 2020.
\newblock Spatial-temporal multi-cue network for continuous sign language recognition.
\newblock In \emph{AAAI}, pages 13009--13016.

\bibitem[{Zhou et~al.(2021{\natexlab{b}})Zhou, Tam, and Lam}]{zhou2021signbert}
Zhenxing Zhou, Vincent~WL Tam, and Edmund~Y Lam. 2021{\natexlab{b}}.
\newblock {SignBERT}: a {BERT}-based deep learning framework for continuous sign language recognition.
\newblock \emph{IEEE Access}, 9:161669--161682.

\bibitem[{Zhou et~al.(2022)Zhou, Tam, and Lam}]{hku.portable.hksl.2022}
Zhenxing Zhou, Vincent~WL Tam, and Edmund~Y Lam. 2022.
\newblock A portable sign language collection and translation platform with smart watches using a {BLSTM}-based multi-feature framework.
\newblock \emph{Micromachines}, 13(2):333.

\bibitem[{Zuo and Mak(2022{\natexlab{a}})}]{zuo2022c2slr}
Ronglai Zuo and Brian Mak. 2022{\natexlab{a}}.
\newblock {C2SLR}: Consistency-enhanced continuous sign language recognition.
\newblock In \emph{CVPR}, pages 5131--5140.

\bibitem[{Zuo and Mak(2022{\natexlab{b}})}]{zuo2022local}
Ronglai Zuo and Brian Mak. 2022{\natexlab{b}}.
\newblock Local context-aware self-attention for continuous sign language recognition.
\newblock In \emph{Interspeech}, pages 4810--4814.

\bibitem[{Zuo et~al.(2023)Zuo, Wei, and Mak}]{zuo2023natural}
Ronglai Zuo, Fangyun Wei, and Brian Mak. 2023.
\newblock Natural language-assisted sign language recognition.
\newblock In \emph{CVPR}.

\bibitem[{Zuo et~al.(2024)Zuo, Wei, and Mak}]{zuo2024towards}
Ronglai Zuo, Fangyun Wei, and Brian Mak. 2024.
\newblock Towards online sign language recognition and translation.
\newblock \emph{arXiv preprint arXiv:2401.05336}.

\end{thebibliography}
